\pdfoutput=1

\documentclass[11pt]{article}

\usepackage[]{acl}

\usepackage{times}
\usepackage{latexsym}
\usepackage{graphicx}
\usepackage{CJKutf8}
\usepackage{multirow}
\usepackage{tabularx}
\usepackage{threeparttable}
\usepackage{afterpage}
\usepackage[T1]{fontenc}

\usepackage[utf8]{inputenc}

\usepackage{microtype}

\title{Exploring Methods for Building Dialects-Mandarin Code-Mixing Corpora: A Case Study in Taiwanese Hokkien}

\author{
Sin-En Lu\textsuperscript{\textdagger}, 
Bo-Han Lu\textsuperscript{\textdagger}, 
Chao-Yi Lu\textsuperscript{\textdaggerdbl},
Richard Tzong-Han Tsai\textsuperscript{\textdagger}\textsuperscript{\textsection}\Thanks{ Corresponding author.}
\smallskip 
\\
\textsuperscript{\textdagger}Department of Computer Science and Information Engineering, 
\\
National Central University, Taiwan
\textsuperscript{\textdaggerdbl}Purdue University, USA
\\
\textsuperscript{\textsection}Center for GIS, Research Center for Humanities and Social Sciences,
\\
Academia Sinica, Taiwan
\\
\texttt{alznn3306@gmail.com, 110522028@cc.ncu.edu.tw}
\\
\texttt{lu940@purdue.edu, thtsai@g.ncu.edu.tw}
}

\begin{document}
\maketitle
\begin{CJK*}{UTF8}{bkai}
\begin{abstract}
In natural language processing (NLP), code-mixing (CM) is a challenging task, especially when the mixed languages include dialects. In Southeast Asian countries such as Singapore, Indonesia, and Malaysia, Hokkien-Mandarin is the most widespread code-mixed language pair among Chinese immigrants, and it is also common in Taiwan. However, dialects such as Hokkien often have a scarcity of resources and the lack of an official writing system, limiting the development of dialect CM research. In this paper, we propose a method to construct a Hokkien-Mandarin CM dataset to mitigate the limitation, overcome the morphological issue under the Sino-Tibetan language family, and offer an efficient Hokkien word segmentation method through a linguistics-based toolkit. Furthermore, we use our proposed dataset and employ transfer learning to train the XLM (cross-lingual language model) for translation tasks. To fit the code-mixing scenario, we adapt XLM slightly. We found that by using linguistic knowledge, rules, and language tags, the model produces good results on CM data translation while maintaining monolingual translation quality.

\end{abstract}

\section{Introduction}
Code-switching or code-mixing (CM), which stands for using more than one language in one conversation or sentence, often occurs in multilingual societies. Because of the rapid development of social media, CM has become more prevalent in the past decade, making it be a new challenge in natural language processing (NLP).

Although Mandarin is the dominant language in Taiwan, Taiwanese Hokkien has nearly as many speakers as Mandarin \cite{TAT_corpus}. Taiwanese tend to mix dialects and Mandarin in daily communication, creating code-mixed languages such as Taiwanese Hokkien-Mandarin or Hakka-Mandarin. Compared with Hokkien, Taiwanese Hokkien integrates Japanese phrases and culture due to historical factors, and gradually evolved into a different dialect. Unlike other CM languages based on recognized writing systems, such as Spanish-English, Hindi-English, or Bahasa Rojak, there were no official writing systems for the dialects in Taiwan until the government built one in the 21st century. Therefore, there is nearly no corpus for Hokkien-Mandarin or other Taiwanese code-mixed languages. Under this circumstance, code-mixing-based NLP tasks (CM tasks) are even more difficult to address in comparison with other monolingual NLP tasks.

The lack of resources \citep{hedderich-etal-2021-survey_LRL}, originating from having no formal writing system for Taiwanese Hokkien in the past, makes it hard to make breakthroughs in dialect-related CM tasks. Researchers are often stuck because of lacking corpus to develop deep learning models. Furthermore, not having a writing system increases the possibility of a language's vanishing \citep{bernard1996language}. Therefore, creating a code-mixed corpus is vital for stepping into the CM-related NLP realm and it also helps to protect the dialects. After creating a code-mixed corpus, more NLP tasks such as machine translation can be developed.

Pre-trained language models have achieved outstanding performance in many NLP tasks. Language models \citep{BERT2019, XLM_lample2019cross,liu2019roberta} have become the mainstream and needful portion in most NLP areas. However, pre-trained language models rely on large-scale corpora, which is a challenge for low-resource languages \cite{hedderich-etal-2021-survey_LRL}. Transfer learning is a possible solution because it uses the knowledge from high-resource tasks to improve performance on the related task. It can reduce the amount of required training data and widely improve the effectiveness when solving low-resource problems, especially in translation task \citep{transferLearningSurvey2009,zoph-etal-2016-transfer,hedderich-etal-2021-survey_LRL,ijcai_RLNMT-survey}. Some research further focused on machine translation, and showed that multilingual models can generalize monolingual inputs to code-switching sentences \cite{johnson-etal-2017-googles,pires-etal-2019-multilingual} without being specifically trained to learn the representations of CM languages.

In this paper, we take Hokkien-Mandarin, a code-mixed language, as the research target because of its large population of speakers in Taiwan. Due to the shortage of Hokkien-Mandarin corpus, we have done several tasks to overcome the low-resource challenge. First, we proposed a method for Hokkien word segmentation via a Mandarin tokenizer toolkit. The method would not be affected by the language morphology and can maintain the syntax structure of Hokkien. It can be seen as the first linguistics-based solution without training a language-specific word segmentation model. Then, we used the Hokkien word segmentation tool to synthesize a Hokkien-Mandarin code-mixed corpus for further use. After that, we proposed a Hokkien-Mandarin cross-lingual language model and achieved good performance on Hokkien-Mandarin CM translation and maintained the monolingual translation result at the same time.

Our main contributions are as follows: (1) We proposed a method of implementing Hokkien word segmentation. (2) We presented a parallel corpus of 76,013 Hokkien-Mandarin CM sentences and 75,150 non-parallel CM data. (3) We built a Hokkien-Mandarin CM translation model through the cross-lingual model.

\section{Background of Taiwanese Hokkien }
Taiwanese Hokkien, also known as Taiwanese, Hokkien, Taigi, Southern Min, or Min-Nan, is a branched-off variety of Southern Min dialects popular in Taiwan. Under the history background \citep{chen2008}, the ability to use Taiwanese Hokkien declines by age \citep{chen2008,TAT_corpus,Tan_2019,Zheng_2007,weko_431_1,Pan_2016,2020_CM_NTU}. Taiwanese Hokkien has always been the most widely spoken dialect in Taiwan, many people can have conversations in both Mandarin and Taiwanese Hokkien. CM between dialects and Mandarin is a common phenomenon in Taiwan. Previous research shows that CM in Taiwan can be divided into fluent and faltering Hokkien CM scenarios, depending on the individual's ability to master the dialect. Therefore, the degree and proportion of using Hokkien and Mandarin vary from person to person, and there is no universal rule or consensus. In the following paragraphs, we will simply use "Hokkien" to represent "Taiwanese Hokkien".

There are two methods to represent the writing system of Taiwanese Hokkien, logograms, and phonography. A logogram is a written character that represents a word or morpheme. In contrast, phonography is an orthography in which the graphemes correspond to the phonemes of the language. The only logogram writing system is \textit{Written Taiwanese Hokkien} (WTH), which is entirely made of Mandarin characters (Hàn-jī). WTH uses the morpheme and meaning of conventional Mandarin characters instead of their phone to create characters for Taiwanese Hokkien. WTH has an official standard for the writing system and is now taught in schools in Taiwan. On the other hand, there are various phonography writing systems such as \textit{POJ (Pe̍h-ōe-jī)}, \textit{Tai-lo} and \textit{Han-Romanization mixed script (Han-lo)}. Table \ref{tab:writing_sys_example} shows examples of different writing systems in Taiwanese Hokkien.
\begin{table*}[ht]
\small
  \begin{tabularx}{\textwidth}{|c|X|}
    \hline
    WTH & 白話字（POJ）是一款用拉丁（羅馬）拼音系統來寫臺灣的語言的書面文字。因為當初是傳教士引入來的,所以也有人共POJ叫做教會羅馬字,或者是簡稱教羅。不而過現代的使用者袂少毋是教徒,教徒嘛真濟袂曉POJ。\\
    \hline
    Tai-lo & Pe̍h-uē-jī (PUJ) sī tsı̍t khuán iōng Latin (Lô-má) phìng-im hē-thóng lâi siá Tâi-uân ê gí-giân ê su-bīn bûn-jī.
    In-uī tong-tshoo sī thuân-kàu-sū ín--jı̍p-lâi ê, sóo-í ia̍h-ū-lâng kā PUJ kiò-tsò Kàu-huē Lô-má-jī, he̍k-tsiá sī kán-tshing Kàu-lô. Put-jî-kò hiān-tāi ê sú-iōng-tsiá bē-tsió m̄-sī kàu-tôo, kàu-tôo mā tsin tsē bē-hiáu PUJ. \\
    \hline
    POJ & Pe̍h-ōe-jī (POJ) sī chı̍t khoán iōng Latin (Lô-má) phèng-im hē-thóng lâi siá Tâi-ôan ê gí-giân ê su-bīn bûn-jī. In-ūi tong-chho͘ sī thôan-kàu-sū ín--jı̍p-lâi ê, só͘-í ia̍h-ū-lâng kā POJ kiò-chò Kàu-hōe Lô-má-jī, he̍k-chiá sī kán-chheng Kàu-lô. Put-jî-kò hiān-tāi ê sú-iōng-chiá bē-chió m̄-sī kàu-tô͘, kàu-tô͘ mā chin chē bē-hiáu POJ. \\
    \hline
    Han-lo & 白話字（POJ）是一款用拉丁（羅馬）拼音系統來寫臺灣ê語言ê書面文字。In-uī當初是傳教士引入來的,所以也有人 kā POJ 叫做教會羅馬字,或者是簡稱 Kàu-lô。 Put-jî-kò 現代 ê 使用者bē-tsió毋是教徒,教徒mā真濟袂曉 POJ。\\
    \hline
  \end{tabularx}
  \caption{Examples of different writing systems in Hokkien }
  \label{tab:writing_sys_example}
\end{table*}
\subsection{Difficulties in Written Taiwanese Hokkien}
To eliminate the problems caused by pronunciation diversity in Hokkien and considering that the government in Taiwan is promoting WTH as the main writing system of Hokkien, we use WTH as the main writing system in this research. However, when using WTH to address the CM tasks in Hokkien and Mandarin, we will face two main problems:  \textit{ambiguous language boundary} and \textit{literary and colloquial readings} problems. 
\paragraph{Ambiguous Word Boundary} is caused by the most important feature of WTH. The WTH writing system uses Mandarin characters to represent the meaning of Hokkien, and new characters are created as supplements. Sharing character space reduces the barrier to learning Hokkien, but it also raises a new problem: \textit{The definition of the language boundary is vague when Mandarin and Hokkien are mixed}. The homophones of Hokkien and Mandarin cannot be clearly distinguished by the text alone. Also, the meanings of shared characters may change or disappear when code-switching occurs. Therefore, preprocessing is needed while addressing CM data.

\paragraph{Literary and Colloquial Readings} refers to various pronunciations of the same character depending on whether it represents a morpheme in the colloquial or literary lexical layers. This phenomenon is widespread in Sinitic languages and has existed for a long time \citep{Yang_2015}. Therefore, literary and colloquial readings is also one of the characteristics of Hokkien. For instance, the word \textit{"八"} \textit{(eight)} is written as "\textit{pat}" in literary readings, and written as "\textit{peh}" in colloquial readings. However, since literary pronunciation is usually established by convention, special cases require additional attention to avoid misunderstandings. Due to this phenomenon, many different processing strategies are required.

\section{Related Work}
\subsection{Code-Mixing}
CM has been a widely-discussed issue for a long time. There is a wide spectrum of opinions on the reasons and motivation of CM occurrence \citep{Mcclure1977AspectsOC,hoffmann1991introduction,lance1970codes,aguirre1985experimental,bokamba1988code,myers1993common,sridhar1980syntax}. To figure out the rules of CM occurrence, research on CM falls essentially into two types: theoretical (also called formal) study and functional study, both proposing different hypotheses and grammatical constraints \citep{timm1975spanish,poplack1980sometimes_EC,pfaff1979constraints,sridhar1980syntax}. In this paper, we only focus on theories that might be related to our research. The Equivalence Constraint \citep{poplack1978syntactic} reports that code-switches tend to occur at points in discourse where the juxtaposition of Language 1 (L1) and Language 2 (L2) elements do not violate a syntactic rule of either language. \citet{poplack1980sometimes_EC} also proffers the Free Morpheme Constraint, which states that the codes in CM language may be switched after any constituent in discourse provided that the constituent is not a bound morpheme. Matrix Language Frame \citep{joshi1982processing,myers1997duelling} defined the dominant language in a CM text as \textit{matrix language}, and other languages are called \textit{inserted languages} or \textit{embedded languages}. All grammar or syntax rules should be under the dominant language. The Functional Head Constraint \citep{di1986government_nhead,belazi1994xbar_nHead} claims that "the language feature of the complement f-selected\footnote{The term F-select, or select following F-selection rule, is a feature selection method. Simply put, a sentence satisfies not only the syntax structure but also the semantics.} by a functional head, like all other relevant features, must match the corresponding feature of that functional head". This means that a language switch between a functional head and its complement does not happen in natural speech. Notice that The Functional Head Constraint should be language-independent.

Both \citet{Shih_su_1993} and \citet{CM_Discourse} agree that nouns have the largest proportion of transformative words, followed by verbs. Also, adhesive words and function words never appear in language switching and are classified as \textit{function units}. The rest are classified as \textit{content units}. Function units cannot be converted alone while the content unit can be freely converted.
In terms of semantics, most conversion words belong to \textit{common expressions} and \textit{common core expressions}. Researchers believe that the reason for language switching is not the lack of vocabulary, but the expression of different social pragmatic functions.
\subsection{Pre-trained Language Models}
BERT \cite{BERT2019} is the first pre-trained language model, which has achieved outstanding performance in the NLP field. Since BERT has made great improvement in the NLP field, using pre-trained language model \cite{liu2019roberta,lewis-etal-2020-bart} has gradually become standard in NLP tasks. Several studies extended the language model to cross-lingual tasks, such as XLM \citep{XLM_lample2019cross}. XLM is a Transformer \citep{Transformer} based architecture that was pre-trained with one of three language modeling objectives: Causal Language Modeling (CLM), Masked Language Modeling (MLM), and Translation Language Modeling (TLM). CLM helps the system to learn the probability of a word when given the previous words in a sentence, which can be seen as a causal language model. MLM can be regarded as the Cloze task, the model would randomly mask the tokens in the sequence, and learn to predict the masked tokens. TLM is a translation language modeling objective for improving cross-lingual pre-training. XLM has achieved state-of-the-art performance on multiple cross-lingual understanding (XLU) benchmarks, and has also obtained significant improvement in both supervised and unsupervised neural machine translation tasks. 
\subsection{NLP task in Code-Mixing}
Neural network models for NLP rely on labeled data for effective training \citep{schuster-etal-2019-cross-lingual}. To deal with CM tasks with neural networks, it is necessary to prepare a large corpus. Some previous CM research collected the corpora from the real world, such as text messages or the internet. Other research also collected them by manually translating monolingual data to CM data \citep{Towards_Translating_Mixed-Code,CM_crawler,lee-wang-2015-emotion,sharma-etal-2016-shallow,banerjee-etal-2018-dataset,singh-etal-2018-twitter,dhar-etal-2018-enabling,chakravarthi-etal-2020-corpus,xiang-etal-2020-sina,srivastava-singh-2020-phinc}. Apart from preparing data manually, a popular strategy for obtaining CM data is through data augmentation.

\citet{pratapa-etal-2018-language} proposed a method to synthesize CM data which is established on the Equivalence Constraint Theory. The researchers designed a computational approach to create a grammatically valid CM corpus by parsing the pair of equivalent sentences and reducing the perplexity of the RNN-based language model through their proposed dataset. Apart from linguistic theory, there is a lot of research focused on generating data through neural-network-based methods, including GAN \citep{goodfellow2014generative}-based method \citep{Chang2019CodeswitchingSG,gao2019code}, a deep generative model \citep{ijcai_VAE_CMdata}, multi-task learning based \citep{winata-etal-2018-code,gupta-etal-2020-semi}, pointer-generator network method \citep{winata-etal-2019-code}, and regarding generating CM corpus as a translation task \citep{gupta-etal-2021-training,gautam-etal-2021-comet}.

\citet{sinha-thakur-2005-machine} is one of the earliest CM translation studies which separated CM translation into three parts. First, identify the language of each word. Then, the recognized noun and adverb phrases in one language are translated into the other language. Finally, translate the language-unified sentences to the final target sentence. \citet{rijhwani2016translating} put forward a similar concept of \citet{sinha-thakur-2005-machine}, concretely defining the task of each step in a CM translation system architecture. Their idea of CM translation had deeply affected the research of CM translation \citep{towards_cmtranslation_social_media,rijhwani2016translating,dhar-etal-2018-enabling,mahata2019code,srivastava-singh-2020-phinc} in the next few years.

\begin{table}[t]
\footnotesize
\centering
\begin{tabular}{|l|l|l|}
\hline
\textbf{Type} & \multicolumn{1}{c|}{\textbf{Data}} & \multicolumn{1}{c|}{\textbf{\#Content}} \\ \hline
Mono.   & Taiwanese songs                    & 30 songs                                \\ \hline
Mono.   & elementary school text books       & 349 articles                            \\ \hline
Mono.   & Hokkien Reading Competition        & 550 articles                            \\ \hline
Mono.   & Subtitles of TV programs           & 126,578 sent.
                        \\ \hline
Para.      & iCorpus                            & 64,110 sent.                        \\ \hline
Para.      & MoE's Dictionary (MoeDict)         & 14,985 sent.                        \\ \hline
\end{tabular}
\caption{Statistics of Hokkien Corpus. \textit{Mono.} refers to monolingual data, and \textit{Para.} refers to parallel data. \textit{Sent.} refers to sentences.}
\label{tab:statistics_of_corpus}
\end{table}

\begin{table*}[ht]
\footnotesize
\begin{tabularx}{\linewidth}{|c|X|X|X|}
\hline
\textbf{English} & Don't strew things all over the ground. & How much do you make a month? &You don't be so serious with him.\\
\hline
\textbf{Mandarin}&東西不要撒得滿地都是&你一個月賺多少錢？&你不要跟他計較\\
\hline
\textbf{ Hokkien}&物件毋通掖甲一四界&你一月日趁偌濟錢？&你毋通佮伊計較\\
\hline
\textbf{Expected}&物件,毋通,掖,甲,一四界&你,一月日,趁,偌濟,錢？&你,毋通,佮,伊,計較\\
\hline
\textbf{Articut}&物件,毋通,掖,甲,一四界&你,一月日,趁,偌濟,錢？&你,毋通,佮,伊,計較\\
\hline
\end{tabularx}
\caption{ Hokkien Sentence Word Segmentation Results in Articut.}
\label{tab:articut_expamle}
\end{table*}

\section {Hokkien-Mandarin CM Dataset}
Hokkien is one of the most popular dialects in Taiwan \cite{kloter2004language,rubinstein2016other}, and switching between Hokkien and Mandarin is very common. However, to the best of our knowledge, there is no code-mixed dataset for Hokkien and Mandarin. Establishing a CM dataset is a challenge we have to overcome.

In our study, we collect two types of data: monolingual data in Mandarin and Hokkien separately, and parallel data in Hokkien-Mandarin. For the Mandarin corpus, we collect the latest Mandarin corpus from Wikipedia. We also gather Taiwan news as a corpus from 2018 to 2019. We collect 2.2 GB of data for training a Mandarin language model. 
For Hokkien and Mandarin parallel corpus, we used iCorpus\footnote{\url{https://github.com/Taiwanese-Corpus/icorpus\_ka1\_han3-ji7}} and the example sentences from MoE's Dictionary of Frequently-Used Taiwan Minnan\footnote{\url{https://github.com/g0v/moedict-webkit}} (MoeDict).

\subsection{Hokkien Dataset}

\label{sec:Hokkien_dataset}
In Hokkien monolingual dataset, our resources contain Taiwanese songs\footnote{\url{https://github.com/Taiwanese-Corpus/Linya-Huang\_2014\_taiwanesecharacters}}, textbooks of elementary school\footnote{\url{https://github.com/Taiwanese-Corpus/kok4hau7-kho3pun2}}, and articles from Hokkien Reading Competition\footnote{\url{https://language110.eduweb.tw/Module/Question/Index.php}}, and the subtitles of Hokkien television program from Chinese Public Television. Note that we only select articles from the Hokkien Reading Competition at the high school level and above, and we excluded data that may contain unrecognized characters or Han-lo script.

For parallel datasets, iCorpus is organized and produced by Academia Sinica, it contains 3,266 news reports from Formosa TV, totaling 64,110 sentences. Including punctuation marks, iCorpus has about 500K Hokkien and 1M Mandarin words. In MoE's Dictionary, there are around 15K example sentences with corresponding Mandarin translations, which were manually created for Hokkien education.

Table \ref{tab:statistics_of_corpus} shows the statistics of the final Hokkien data. All data are from open-source resources or public government data, collected and organized by the author, and are only used for academic research.

\subsection{Hokkien Word Segmentation}
Word Segmentation is usually an important step when processing Sinitic languages. However, there is no open-source word segmentation tool for Hokkien, so we need to develop one before other tasks. The most typical word order in Hokkien is \textit{Subject}, \textit{Verb}, and \textit{Object}, which is also the same as that in Mandarin. But there are many sentence patterns with more complicated structures and diverse grammar rules in Hokkien \citep{Tang_1999}. To synthesize Hokkien-Mandarin CM sentences under constraints, it is important to parse the structure of Hokkien sentences precisely. Due to the syntactic complexity of Hokkien, the majority of Mandarin NLP toolkits sometimes provide unexpected results when used with Hokkien sentences. The ability to address unknown words is below our expectations, resulting in losing word boundaries. Also, we may lose the part-of-speech (POS) or syntax information while parsing the sentences. It is difficult to synthesize the CM sentences according to incorrect word boundaries or syntactic parsing results. Our experiment reveals some issues while using BERT-based Mandarin tokenizers, please refer to \ref{sec:issue_of_CKIP} for more details.

To capture word boundaries and parse Hokkien syntax, another Mandarin tokenizer, Articut \citep{wang_articut_2021}, is our solution. There are two reasons why we believe applying Articut to implement Hokkien word segmentation is a potential solution. First, both Hokkien and Mandarin belong to Sino-Tibetan Family. The positions of functional heads in the same language family are almost the same \citep{Tang_1999}. Therefore, the syntax of Hokkien and Mandarin are similar (can even be regarded as the same) in linguistics. Second, the working principle of Articut is the X-bar theory \citep{xbar1970Chomsky}, which makes it possibly the only tokenizer designed based on linguistics.

According to \citet{xbar1970Chomsky}, the X-bar stands for that every phrase in every sentence in every language is arranged in the same way. Each phrase has a head and may include other phrases in the complement or specifier position. X-bar embodies two central principles, Headedness Principle and Binarity Principle. In the Headedness Principle, every phrase has a head. In the binarity principle, every node branches into two different nodes. X-bar relies on these functional heads to check the POS of each word forward and backward. Through the binary structure of the X-bar, Articut can calculate the vocabulary boundary and determine their POS at the same time. As a result, in Hokkien word segmentation, we only need to adjust some "internal order" of morphology. That is, Articut can be adapted to Hokkien, which is mainly designed for Mandarin. Moreover, the issue of morphology can be solved by the custom dictionary provided by Articut.

Second, the Functional Head Constraint proposed by \citet{belazi1994xbar_nHead} is actually based on the assumption of X-bar theory. The constraint follows \citet{chomsky1993minimalist}, assuming that f-selection\cite{belazi1994code}, a special relationship between a functional head and its complement, is one member of a set of feature-checking processes. In a nutshell, when every noun has a proper position and each functional head works smoothly in a sentence, without exception, a complete syntax tree can be generated.

Therefore, when Articut is processing an input phrase, it checks whether the input satisfies the X-bar theory so that a complete syntactic tree can be generated. After the syntactic tree be successfully created, it signifies that the feature-checking operations have been finished once and the functional head in the sentence can be grasped successfully. Hence we can grab the word boundary and the POS tagging in Hokkien sentences.

We apply MoE's Dictionary of Frequently-Used Taiwan Minnan\footnote{\url{https://twblg.dict.edu.tw/holodict\_new/}} as the custom dictionary of Articut to implement Hokkien word segmentation. As shown in Table \ref{tab:articut_expamle}, Articut can parse sentences correctly and is not affected by unknown words or morphology. The result verified our conjecture. Through X-bar and a custom dictionary, Articut can correctly identify word boundaries. 

\subsection{Synthesis of Code-Mixed Corpus}
\label{sec:dataset}



After collecting the corpora, we first normalize the data through the rule-based method to deal with the literary and colloquial reading issue in the Hokkien corpora. We convert words with this issue into Mandarin characters. Furthermore, there are a lot of Tai-lo words that are regarded as noise in our data, so it is necessary to convert them into pure Written Taiwanese Hokkien. The next step is to synthesize a code-mixed corpus. Similar to \citet{pratapa-etal-2018-language}, we synthesize the Hokkien-Mandarin CM dataset based on the matrix language frame, the equivalence constraint, and the functional head constraint. In our work, we defined Hokkien as the matrix language and Mandarin as the embedded language. 

We then applied the equivalence constraint and the functional head constraint to the sentences from the 2 parallel corpora, iCorpus and MoeDict. We first build a Hokkien-Mandarin dictionary based on MoE's Dictionary of frequently used Taiwan Hokkien, and then parse the Hokkien sentence using Articut to get word boundaries and POS tags. Finally, we switch the word to the corresponding Mandarin word according to the dictionary. Under the functional head constraint and the equivalence constraint, the language switch point in our synthetic progress is based on several previous studies \citep{wu-etal-2011-YZU,Shih_su_1993,CM_Discourse,chen1989}.
The switch point rules are as follows: (1) If \textbf{Head Noun} appears in the sentence, switch the \textbf{Head Noun}. (2) Switch \textbf{Idioms}, but keep the common sayings and proverbs. (3) Switch all \textbf{Person} and \textbf{Location} in sentence. (4) Switch \textbf{Noun Phrase} and \textbf{Verb Phrase} in the sentence. (5) Switch the \textbf{Noun} after \textbf{Preposition}. Lastly, to distinguish the Hokkien and Mandarin, we add \textbf{\_@} after each Hokkien character.

We produce a total of four datasets, iCorpus-CM, iCorpus-CMDA\footnote{DA means Data Augmentation.}, MoeDict-CM and MoeDict-CMDA\footnote{\url{https://github.com/alznn/Taiwanese-Hokkien_Mandarin_CM_Dataset}}, by two different methods. The first type, iCorpus-CM and MoeDict-CM, means that sentences match all the above rules of switching points and all the constraints, and all the words are precisely translated. The second type, iCorpus-CMDA and MoeDict-CMDA, means that sentences match all of the rules of switching points but not all the constraints, sentences might contain ambiguously translated words. The statistics of four datasets are shown in Table \ref{tab:statistics_CM}. The CM dataset complexity: Switch Point Fraction (SPF) \citep{pratapa-etal-2018-language} and Code-Mixing Index (CMI) \citep{Gambck2014OnMT,gamback-das-2016-comparing}, are also reported in Table \ref{tab:statistics_CM}. For the summary of all datasets in our work, please refer to Table \ref{tab:Summary_of_all_Corpus}. And the examples of CM sentences in our data are shown in Table \ref{tab:CM_expamle}. 

\begin{table}[t]
\small
  \centering
  \begin{tabular}{|c|c|c|c|c|}
    \hline
    Corpus & Sent. & Symbol & CMI & SPF \\
     & Nums. & Coverage & &  \\
    \hline
    iCorpus-CM & 61,690 & 0.1813 & 0.571 & 0.301 \\
    \hline
    iCorpus-CMDA & 63,604 & 0.1525 & 0.497 & 0.306 \\
    \hline
    MoeDict-CM & 12,409 & 0.1847 & 0.483 & 0.267 \\
    \hline
    MoeDict-CMDA & 13,460 & 0.1907 & 0.374 & 0.229 \\
    \hline
  \end{tabular}
  \caption{Statistics of CM Datasets.}
  \label{tab:statistics_CM}
\end{table}


\begin{table*}[ht]
\small
\begin{tabularx}{\linewidth}{|c|c|X|}
    \hline
    Type & Language & Dataset\\
    \hline
    Monolingual & Hokkien & Taiwanese songs, elementary school textbooks, Hokkien Reading Competition, subtitles of Hokkien television program\\
    \hline
    Monolingual & Mandarin & Mandarin Wiki, News \\
    \hline
    Parallel & Hokkien-Mandarin & iCorpus, MoE's Dictionary\\
    \hline
     Parallel & CM-Mandarin & iCorpus-CM, iCorpus-CMDA, MoeDict-CM, MoeDict-CMDA,\\
    \hline
  \end{tabularx}
  \caption{Summary of all Corpora}
  \label{tab:Summary_of_all_Corpus}
\end{table*}

\begin{table*}[ht]
\small
\begin{tabularx}{\linewidth}{|c|X|}
\hline
\textbf{Corpus}& Example\\
\hline
\textbf{English} & Announced at 11 o'clock in the evening Eastern time on the 4th.\\
\hline
\textbf{Mandarin}&在美東時間四日深夜十一時宣布\\
\hline
\textbf{iCorpus}&佇美東時間四號深夜十一點宣布\\
\hline
\textbf{iCorpus-CM}&佇\_@ \textbf{美 東 時 間} 四\_@ 號\_@ \textbf{子 夜} 十\_@ 一\_@ 點\_@ 宣\_@ 布\_@\\
\hline
\textbf{iCorpus-CMDA}&佇\_@ \textbf{美\_@ 東\_@ 時\_@ 間\_@} 四\_@ 號\_@ \textbf{深 夜} 十\_@ 一\_@ 點\_@ 宣\_@ 布\_@\\
\hline
\hline
\textbf{English} & This doesn't work. That doesn't work. You have so many opinions.\\
\hline
\textbf{Mandarin}&這個不行,那個不可以,你的意見真多。\\
\hline
\textbf{MoeDict}&這\begin{CJK}{UTF8}{gbsn}个\end{CJK}袂使,彼\begin{CJK}{UTF8}{gbsn}个\end{CJK}毋通,全你的意見了了。\\
\hline
\textbf{MoeDict-CM}& 這\_@ \begin{CJK}{UTF8}{gbsn}个\_@\end{CJK} \textbf{不 可} , 彼\_@ \begin{CJK}{UTF8}{gbsn}个\_@\end{CJK} 毋\_@ 通\_@ , 全\_@ 你\_@ 的\_@ \textbf{意 見} 了\_@ 了\_@ 。\\
\hline
\textbf{MoeDict-CMDA}&這\_@ \begin{CJK}{UTF8}{gbsn}个\_@\end{CJK} \textbf{袂\_@ 使\_@} , 彼\_@ \begin{CJK}{UTF8}{gbsn}个\_@\end{CJK} 毋\_@ 通\_@ , 全\_@ 你\_@ 的\_@ \textbf{意 見} 了\_@ 了\_@ 。\\
\hline
\end{tabularx}
\caption{ CM sentence example in each corpus.}
\label{tab:CM_expamle}
\end{table*}

\begin{table*}[t]
  \centering
  \footnotesize
  \begin{tabularx}{\textwidth}{c|cccc|cccc}
    \hline
    \multirow{3}{*}{Annotator} & \multicolumn{4}{c|}{iCorpus ( Sent:1879)} & \multicolumn{4}{c}{MoeDict ( Sent:179 )} \\
    & Colloquialism & Intelligibility & Coherence & Total & Colloquialism & Intelligibility & Coherence & Total\\
    \hline
    A & 2.351 & 2.463 & 2.422 & 3.608 & 2.268 & 2.503 & 2.307 & 3.374 \\
    \hline
    B & 2.353 & 2.470 & 2.560 & 3.949 & 2.223 & 2.358 & 2.542 & 3.810 \\
    \hline
    C & 1.884 & 2.494 & 2.767 & 3.537 & 1.502 & 2.564 & 2.721 & 3.134 \\
    \hline
    Avg.& 2.20 & 2.48 & 2.58 & 3.70 & 2.00 & 2.47 & 2.52 & 3.44\\
    \hline
  \end{tabularx}
  \caption{Result of human scoring in CM dataset. Annotator A, B, and C represent elders,  the middle-aged population, and youngsters, respectively.}
  \label{tab:human_scoring}
\end{table*}
\section{Data Quality}

\label{sec:Human Evalutaion}
\subsection{Human Scoring}
For human evaluation, we hired three annotators. One of them holds the intermediate level of Hokkien language proficiency certification from the Ministry of Education\footnote{\url{https://blgjts.moe.edu.tw/tmt/index.php}} and the advanced level of Hokkien accreditation from National Cheng-Kung University\footnote{\url{https://ctlt.twl.ncku.edu.tw/gtpt/index.html}}. The other has the junior-level certification of Hokkien language proficiency from the Ministry of Education. The last one has no certification. The annotators are from three different generations of Taiwanese speakers, elders, middle-aged, and youngsters. We will use annotators A, B, and C to represent each annotator, respectively. All annotators are anonymous and they do not know each other. The annotators are clearly informed of the purpose and use of the entire experiment before the annotating. Each annotator spends 30 hours scoring the data. The annotating costs total \$480 US dollar.

The annotators were asked to read 1,879 CM sentences sampled from iCorpus-CM and 179 CM sentences sampled from MoeDict-CM in two phases. In the first phase, the annotators need to score the CM sentence on a scale of 1 (very poor) to 5 (excellent). Our grading criteria are based on the \citet{khanuja2020new} and we widen the score interval. The instructions to the annotators are as follows:
\begin{enumerate}
\item{\textbf{Very poor}: The sentence is unreasonable, extremely unnatural, and does not exist in daily life.}
\item{\textbf{Poor}: The sentence is reasonable but slightly unnatural. It takes time to comprehend the sentence's meaning. The sentence structure barely exists.}
\item{\textbf{Fair}: The sentence is reasonable and natural. The sentence is fairly easy to understand and may appear in daily conversations.}
\item{\textbf{Good}: The sentence is reasonable and natural. It is well structured and expected to appear in daily conversations.}
\item{\textbf{Excellent}: The sentence is reasonable and fluent. It is well-structured and often used in daily conversations.}
\end{enumerate}

Given that CM behaviors vary from person to person, the scores might be affected by the annotator's life experience. We further ask the annotators to evaluate the sentence in three aspects, colloquialism, coherence, and intelligibility. The metrics were inspired
from \citet{banerjee-etal-2018-dataset} and have slight modifications on the definition to fit the characteristics of our dataset. The annotators need to follow our instructions and rate each metric on a scale of 1 (poor) to 3 (good). The metrics are described as follows:
\begin{enumerate}
  \item{\textbf{Colloquialism}: Check whether the CM sentence is colloquial enough that people may use it in daily life.}
  \item{\textbf{Intelligibility}: Check whether the words used in the switching point are correct, including POS and meaning. Ensure all sentences are easy to understand.}
  \item{\textbf{Coherence}: Check whether the language switch point is reasonable, the CM sentence was smooth enough and not forced.}
\end{enumerate}
The results are shown in Table \ref{tab:human_scoring}. According to Table \ref{tab:human_scoring}, it is clear that the evaluation of colloquialism is quite different from the annotator C and others. After investigation, we found that it is because many vocabularies were written in WTH, which the annotator C was not sure how to pronounce. Therefore the annotator C gave a lower score for colloquialism.

We suppose this reflects that even though WTH is designed according to Mandarin characters, it is still hard for understanding the relationship between the spoken language system and the writing system.


\subsection{Inter-rater score}
We also calculated the kappa value \citep{kappa} between our annotators. Similar to \citet{dhar-etal-2018-enabling}, we randomly selected 100 sentences from our parallel corpus and then requested one of the annotators to translate them into CM sentences to evaluate the reliability of our human evaluation. The other two annotators were assigned to rate the translated sentences into three categories, Totally Agree, Fair Agree, and Disagree. Then, we consider the classification of Agree label as True and Disagree label as False. Finally, we calculate the kappa value by using these labels. The final kappa value is about 0.740. All data we used and created are open source and for academic purposes only.

\begin{figure*}[t]
  \centering
  \includegraphics[width=380pt]{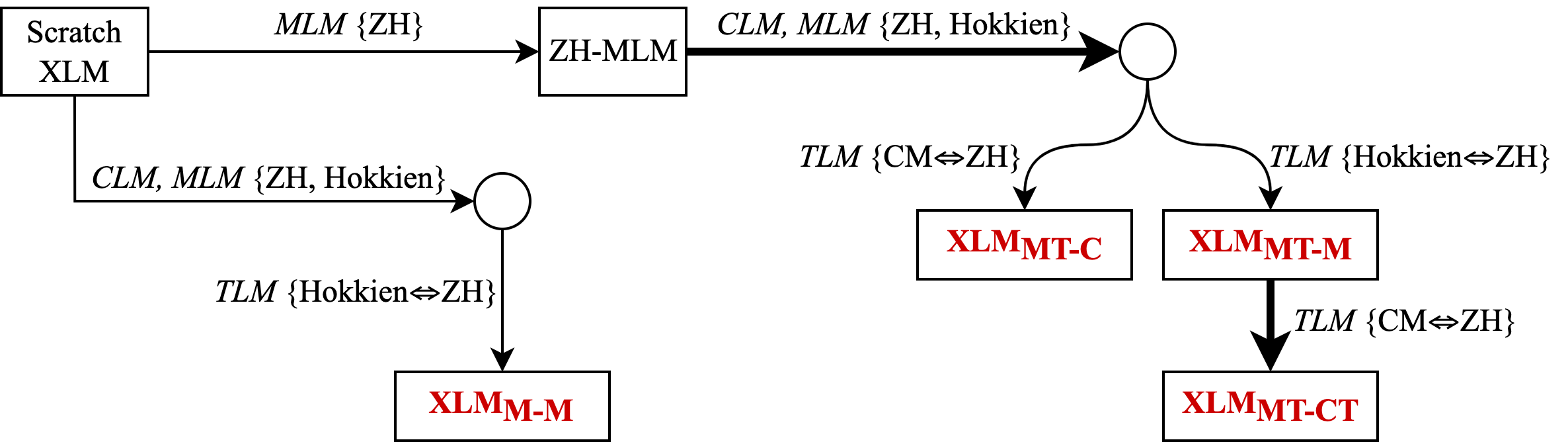}
  \caption{ The training process of the model. In XLM\(_{\mathrm{X-Y}}\), we use X for CLM, MLM stage and Y for TLM stage. M: train on monolingual data only. T: utilize transfer learning. C: the training resource contains CM data. (The processes using transfer learning are indicated by bold arrow symbols.) }
  \label{fig:model_training_process}
\end{figure*}
\section{Hokkien Language Model and Translation System}
\label{sec:assump}
Once we synthesize the code-mixing corpus, we can intuitively apply the data to various NLP tasks. In this study, we chose to apply our CM dataset to the translation task. The main advantage of developing a translation model is that we can translate CM sentences to monolingual sentences and use them as input data for other NLP applications without retraining CM-based NLP models. 

Previous research \cite{johnson-etal-2017-googles,pires-etal-2019-multilingual} has pointed out that multilingual models can generalize well on CM data. We believe that since bilingual people can use knowledge of one language to aid the learning of the other, they can identify the language to which the vocabulary belongs in a code-mixed sentence. Also, understand the meaning of the sentence without learning the sentence structure and grammar of the code-mixed language.

Considering the size of our corpus, we use XLM as our multilingual model architecture. Following XLM, we build a vocabulary dictionary at the character level, including all Mandarin and Hokkien characters, Roman numerals, and English letters, for a total of 26,780 characters. Same as our CM dataset, we distinguish Mandarin and Hokkien characters by appending the \textbf{\_@} symbol when building the vocabulary set. We also keep the same special token as XLM. At last, we applied two methods to XLM to verify our hypothesis, dynamic language identification (DLI) mechanism and transfer learning.

\paragraph{Dynamic Language Identification} Compared with the pre-defined language embedding input in XLM, the DLI mechanism can dynamically distinguish whether the language is Hokkien or Mandarin through the \_@ symbol, and assign the corresponding language embedding to each word. Not only does it suit CM data scenarios where sentences consist of multiple languages, but it also allows XLM to detect which language each word belongs to.

\paragraph{Transfer Learning} To figure out if the model can leverage monolingual knowledge to process CM sentences well, we attempt to apply transfer learning to three training objectives (CLM, MLM, TLM) used in XLM. We trained the XLM model from scratch using only ZH-Hokkien corpus as our baseline model \(\mathrm{XLM_{M-M}}\). The training process for other models is divided into three stages. 1) We perform an MLM pre-training strategy on the ZH corpus, called ZH-MLM. 2) We continue pre-training the model from the first stage on the both ZH and Hokkien corpus, using CLM and MLM. 3) Once the model in the second stage converges, we continue training the TLM. We
train the TLM model using CM-ZH and Hokkien-ZH parallel data, called \(\mathrm{XLM_{MT-C}}\) and \(\mathrm{XLM_{MT-M}}\) respectively. At last, we utilize the parameters from \(\mathrm{XLM_{MT-M}}\) to continue training TLM on the CM-ZH corpus, called \(\mathrm{XLM_{MT-CT}}\). The process of how we obtain these models is also shown in Figure \ref{fig:model_training_process}.

\section{Experiment}
For the dataset, taking the diversity in CM habits into consideration, we sample some data from the Hokkien corpus and synthesize them into the CM data randomly. We shuffle all the sentences and split them into training, validation, and test sets in a ratio of 8:1:1 in each parallel corpus. Moreover, a few parts of our Hokkien corpus which can be found corresponding to Mandarin translation were reserved for testing. 
There are 823 sentences in total, and we use them to synthesize the CM preserved assessment dataset (PAD) by the same method. PAD is not used in any pre-training stage of the XLM model. The CMI and SPF in PAD data are 0.537 and 0.3, respectively. Our evaluation metrics are BLEU \citep{papineni-etal-2002-bleu} and BERT-Score \citep{bert-score}. All experiments were done three times and we report the average score. In the model configuration, we set the dimension of embedding to 768. The rest of the configuration follows the vanilla XLM. All models are trained on NVIDIA 3090 GPU. It took 7 days to train ZH-MLM and about 1 to 3 days for the rest of the models.

Our experiment results show that using DLI brings a positive influence on most configurations of the model. Furthermore, transfer learning in the CLM and MLM stages significantly improves performance. It demonstrates that continuous training with our CM data enables monolingual language models to provide better performance when applied to CM translation tasks. For more details and discussions in our experiment, please refer to \ref{sec:experiments}.
\section{Conclusion}
In this paper, we introduce the Hokkien background and the CM phenomenon between dialects and Mandarin in Taiwan. We proposed a Hokkien-Mandarin CM dataset based on the linguist theory and the Hokkien grammar rules. We proposed a solution to the Hokkien word segmentation through a linguistics-based toolkit, Articut. Based on the X-bar theory, we can avoid the negative impact of morphology on the Sino-Tibetan languages. We simply modify the language embedding mechanism and use transfer learning in the XLM model, which performs well on both CM and monolingual translation. We again prove that under a fully trained language model and well-defined language identity, the cross-lingual model can generalize the knowledge to a CM sentence without special training.
We also verify the feasibility of linguistic-based background knowledge as a low-resource language solution. After developing the CM corpora and translation systems, it can be further extended to any existing monolingual task. Meanwhile, we can use them to generate speech data, and train a CM speech recognition model. 

\section*{Limitations}

Our research is designed for the Sino-Tibetan language family. However, the language features may not be generalized to other language families well. Furthermore, in our case, Hokkien has officially defined Mandarin characters. Without officially defined characters, it might be difficult to eliminate the differences between writing systems and create useful datasets. Finally, during data construction, we can directly mark the language, so we can assume language identification has 100\% accuracy in the translation model. Without a good language identification system, the performance of the translation model might be affected.
\section*{Acknowledgements}
The authors wish to thank prof. Iûⁿ\ Ún-giân and all previous giants who devote their lives to preserving, studying, and promoting Taiwanese dialects. Special thanks to Wang, Wen-jet, CEO of Droidtown Linguistic Tech. Co., Ltd., who selflessly imparts linguistic knowledge to the authors and shares the design context of Articut. Discuss using Articut to implement Taiwanese Hokkien word segmentation with authors. We want to express our sincere gratitude to Dr. Ching-Ching Lu, and all members of the IISR research team, especially Dr. Yu-Chun Wang, Chun-Kai Wu, Ka-Ming Wong, and In-Zu Gi, for the generous assistance, discussion, and advisement they provided for this paper. This research was supported in part by the National Science and Technology Council under grants 109-2221-E-008 -058 -MY3 and 109-2221-E-008 -058 -MY3.
\bibliography{anthology,custom}
\bibliographystyle{acl_natbib}

\appendix

\section{Appendix}
\label{sec:appendix}

\subsection{Problems of using Mandarin Toolkit}
\label{sec:issue_of_CKIP}
CKIP\footnote{\url{https://github.com/ckiplab/ckiptagger}} is one of the Mandarin NLP toolkits based on BERT. It can be seen as the most robust Mandarin tokenizer toolkit. Our experiment reveals that CKIP may provide unexpected results in Hokkien sentence word segmentation due to the different grammar structure between Hokkien and Mandarin. Some examples of tokenization results are shown in Table \ref{tab:CKIP_expamle_good} and Table \ref{tab:CKIP_expamle_worse}. 

In Table \ref{tab:CKIP_expamle_good}, the words in sentence P1 share the same characters and meanings between WTH and Mandarin. Most words in sentences P2 and P3 share the same characters and meanings. As for the words with different characters, such as \textit{"毋通"} corresponding to \textit{"不要"} in P2, or \textit{"目尾"} corresponding to \textit{"眼角"}, \textit{"皺痕"} corresponding to \textit{"皺紋"} in P3, they have similar meanings ( \textit{"毋"}/\textit{"不"}, \textit{"目"}/\textit{"眼"}, \textit{"皺痕"}/\textit{"皺紋"}) in both languages either. Therefore, the word segmentation results of these sentences are in our expectations. We call them positive cases.

\begin{table*}[ht]
\small
\begin{tabularx}{\linewidth}{|c|X|X|X|}
\hline
\textbf{Sent. Number}& P1 & P2 & P3 \\
\hline
\textbf{English}&Being a professor is my lifelong wish.&You, don't live a life of debauchery.&I started to have wrinkles in the corners of my eyes.\\
\hline
\textbf{Mandarin}&做教授是我一生的願望。&你不要放蕩過一生。&我的眼角開始有皺紋了。\\
\hline
\textbf{Hokkien}&做教授是我一生的願望。&你毋通放蕩過一生。&我的目尾開始有皺痕矣。\\
\hline
\textbf{Expected}&做,教授,是,我,一生,的,願望,。&你,毋通,放蕩,過,一生,。&我,的,目尾,開始,有,皺痕,矣,。\\
\hline
\textbf{CKIP-BERT}&做,教授,是,我,一生,的,願望,。&你,毋通,放蕩,過,一生,。&我,的,目尾,開始,有,皺痕,矣,。\\
\hline
\textbf{CKIP-ALBERT}&做,教授,是,我,一生,的,願望,。&你,毋通,放蕩,過,一生,。&我,的,目尾,開始,有,皺痕,矣,。\\
\hline
\end{tabularx}
\caption{Positive cases of Hokkien Sentence Word Segmentation in CKIP.}
\label{tab:CKIP_expamle_good}
\end{table*}

\begin{table*}[ht]
\small
\begin{tabularx}{\linewidth}{|c|X|X|X|}
\hline
\textbf{Sent. Number}& N1 & N2 & N3\\
\hline
\textbf{English} & Don't strew things all over the ground. & How much do you make a month? &You don't be so serious with her.\\
\hline
\textbf{Mandarin}&東西不要撒得滿地都是&你一個月賺多少錢？&你不要跟他計較\\
\hline
\textbf{Hokkien}&物件毋通掖甲一四界&你一月日趁偌濟錢？&你毋通佮伊計較\\
\hline
\textbf{Expected}&物件,毋通,掖,甲,一四界,。&你,一月日,趁,偌濟,錢,？&你,毋通,佮,伊,計較,。\\
\hline
\textbf{CKIP-BERT}&物件,毋,通,掖,甲,一,四,界,。&你,一,月,日,趁,偌濟,錢,？&你,毋,通,佮伊,計較,。\\
\hline
\textbf{CKIP-ALBERT}&物件,毋,通,掖甲一四,界,&你,一月日,趁,偌,濟錢,&你,毋,通,佮伊,計較,。\\
\hline
\end{tabularx}
\caption{Negative cases of Hokkien Sentence Word Segmentation in CKIP.}
\label{tab:CKIP_expamle_worse}
\end{table*}

\begin{table*}[t]
\footnotesize
  \centering
  \begin{tabularx}{\textwidth}{c|c|cc|cccccc}
    \hline
    \multirow{3}{*}{Model} & \multirow{3}{*}{Setting} &\multicolumn{2}{c|}{PARAMS}& \multicolumn{6}{c}{Test}\\
    & & \multirow{2}{*}{Dim} & Batch & zh & zh & min & min & min\_zh & min\_zh \\
    & &  & Size & mlm\_ppl & mlm\_acc & mlm\_ppl & mlm\_acc & mlm\_ppl & mlm\_acc\\
    \hline
    XLM Scratch & ZH MLM &768 & 32 & 3.818 & 68.949 & - & - & 3.818 & 68.949\\
    \hline
    XLM Scratch & CLM, MLM & 768 & 64 & 9.796 & 55.418 & 8.518 & 55.293 & 1.741 & 88.7494 \\
    \hline
    XLM Transfer & CLM, MLM & 768 & 64 & 4.794 & 65.048 & 5.948 & 62.522 & 1.391 & 92.560 \\
    \hline
  \end{tabularx}
  \caption{Language model in XLM.}
  \label{tab:XLM_LM}
\end{table*}

Table \ref{tab:CKIP_expamle_worse} shows the less ideal examples. The examples N1 and N2 are the word segment results of Hokkien sentences. Each word in these sentences, from the structure of the word to the word itself, has a different meaning in Mandarin. As the phrases \textit{"一四界"}, \textit{"一月日"}, and \textit{"偌濟"} are used in Hokkien only. Also, it is difficult to associate the meanings of individual characters with the meaning of the phrases. Though the phrases \textit{"物件"}, \textit{"掖"}, and \textit{"趁"} exist in Mandarin, the meanings in the two languages are very different. Therefore, the results in N1 and N2 are worse than we expected. 

N3 displays the word segment outcome when a Hokkien sentence follows the same grammar structure as a Mandarin sentence but contains any newly created words. Term \textit{"佮"} is a new word designed for Hokkien, and would not appear in Mandarin corpora, which implies it is an unknown word to a BERT-base tokenizer. Compared with the P2 in Table \ref{tab:CKIP_expamle_good}, N3 consists of the same sentence structure and overlap word \textit{"毋通"}. The remaining words have the same meaning and vocabulary in Hokkien and Mandarin. However, CKIP couldn't parser the \textit{"毋通"} and other words correctly.

It can be observed from the results that the BERT-based model tends to process unintelligible vocabulary in a character-base manner. We speculate the reason is that BERT uses a character-based method to split the Mandarin characters. And most BERT-based toolkits rely on "continuous distribution of context relation in high-dimensional vector spaces" to implement word segmentation or other semantic analysis, which brings too much noise while applying these toolkits on Hokkien sentences.

In CKIP, the ability to address unknown words is lower than we expected. BERT-based processing method will make it lose word boundaries and the POS or syntax information while parsing the sentence. So in this case, it would be difficult to synthesize the CM sentences based on the word boundary or syntactic parsing results.

\subsection{Model Discussion}
\label{sec:experiments}
\subsubsection{Hokkien language model on BERT}
\label{sec:Hokkien_LM}
In this section, we will explain how to transfer a Mandarin BERT model to train a Hokkien language model. 
The main reason we need to train a Hokkien language model is that although most of the characters in Written Taiwanese Hokkien and Mandarin are the same, the grammar and meaning are often different. They still need to be regarded as two different languages. Pre-trained language models have become the standard step for NLP tasks today and rely on a large corpus \cite{hedderich-etal-2021-survey_LRL}. This implies that training a Hokkien Language Model from scratch is unrealistic. Since Written Taiwanese Hokkien and Mandarin share some characters, transfer learning can be applied to solve the above-mentioned problem. Using a pre-trained Mandarin language model and transferring the parameters to a Hokkien language model could be a potential solution.

There are about 800 characters in Written Taiwanese Hokkien that do not exist in Mandarin. To pre-train a Hokkien language model, we first replace the special tokens and rarely used characters in vanilla BERT's vocabulary dictionary with these newly created Hokkien words. The examples of the replaced vocabulary set are shown in Table \ref{tab:unused_word_replacement}. Second, we set a higher priority to mask these new words forcing the model to learn these words during the MLM step. Then, we continue training a BERT-based language model with Mandarin model parameters on the monolingual Hokkien corpus.

We trained our model with 500k steps, 16 batch size, 256 max sequence length, 3000 warm-up steps, and $2\mathrm{e}{-5}$ learning rate. The Hokkien language model achieves over 78\% accuracy on mask word prediction, and the loss function is reduced to $2.23\mathrm{e}{-3}$. The examples of the Hokkien language model predicting results are shown in Table \ref{tab:special_predict}. 

Note that example numbers 1, 2, 6, and 10 are newly created words in Hokkien and the language model can predict them correctly. In the model configuration, we set the dimension embedding to 768. The rest of the configuration follows that of the vanilla XLM. All models are trained on NVIDIA 2080 GPU. It took 12 hours to train the Hokkien LM.

\begin{table}[t]
\small
  \centering
  \begin{tabular}{|c|c|c|}
    \hline
    index & select word & map word\\
    \hline
    3 & [unused2] & 佮\\
    \hline
    4 & [unused3] & \begin{CJK}{UTF8}{gbsn}个\end{CJK}\\
    \hline
    5 & [unused4] & 仝\\
    \hline
    6 & [unused5] & 囥\\
    \hline
    7 & [unused6] & 紲\\
    \hline
    8 & [unused7] & 蹛\\
    \hline
    9 & [unused8] & 爿\\
    \hline
    10 & [unused9] & 徛\\
    \hline
    11 & [unused10] & 翕\\
    \hline
  \end{tabular}
  \caption{Example of replacing an unused word in the BERT vocabulary.}
  \label{tab:unused_word_replacement}
\end{table}

\begin{table}[t]
\small
\centering
\begin{tabular}{|c|c|c|}
\hline
num. & answer & predict\\ 
\hline
1 & \textbf{佮} & \textbf{佮}\\ 
\hline
2 & \textbf{仝} & \textbf{仝}\\
 \hline
3 & 支 & 支\\ 
 \hline
4 & [PAD] & 的\\ 
 \hline
5 & 保 & 保\\ 
 \hline
6 & \textbf{佮} & \textbf{佮}\\
 \hline
7 & 歇 & 歇\\ 
 \hline
8 & 的 & 暗\\ 
 \hline
9 & 一 & 一\\ 
 \hline
10& \textbf{爿} & \textbf{爿}\\
\hline
\end{tabular}
\caption{Example of \texttt{[MASK]} token predicting results.}
\label{tab:special_predict}
\end{table}
\begin{table*}[t]
  \centering
  \footnotesize
  \begin{tabular}{c|c|cc|cc|ccccc}
    \hline
    \multirow{3}{*}{Num} &\multirow{3}{*}{Name} & \multicolumn{2}{c|}{Config.} & \multicolumn{2}{c|}{Testset} & \multicolumn{5}{c}{PAD} \\
    & & \multirow{2}{*}{AE} & \multirow{2}{*}{DLI} &  \multirow{2}{*}{Acc} & \multirow{2}{*}{BLEU} & Mono & CM & \multicolumn{3}{c}{CM BERT-Score}\\
    & &  &  &  &  & BLEU & BLEU & Precision & Recall & F1 \\
    \hline
    \hline
    0 &$\mathrm{XLM_{M-M}}$ & + & + & 90.898 & 75.46 & 51.42 & 54.87 & 87.952 & 89.334 & 88.585 \\
    1 &$\mathrm{XLM_{M-M}}$ & + & - & 91.149 & 75.85 & 50.96 & 49.23 & 87.911 & 89.203 & 88.503 \\
    \hline
    2 &$\mathrm{XLM_{M-M}}$ & - & + & 90.841 & 75.37 & 52.21 & 49.54 & 87.089 & 87.755 & 87.387 \\
    3 &$\mathrm{XLM_{M-M}}$ & - & - & 90.736 & 75.16 & 49.52 & 47.56 & 86.357 & 87.410 & 86.847 \\
    \hline
    \hline
    4 & $\mathrm{XLM_{MT-M}}$ & + & + & 91.887 & 83.15 & 62.59 & 62.11 & 90.633 & 91.645 & 91.097 \\
    5 & $\mathrm{XLM_{MT-M}}$ & + & - & 91.796 & 83.03 & 62.21 & 61.87 & 90.122 & 91.455 & 90.742 \\
    \hline
    6 & $\mathrm{XLM_{MT-M}}$ & - & + & 91.440 & 82.54 & 61.24 & 60.25 & 91.234 & 91.033 & 91.105 \\
    7 & $\mathrm{XLM_{MT-M}}$ & - & - & 91.553 & 82.87 & 60.05 & 57.44 & 89.823 & 89.752 & 89.766 \\
    \hline
    \hline
    8 & $\mathrm{XLM_{MT-C}}$ & + & + & 91.964 & 82.92 & 60.38 & 62.86 & 90.670 & 91.678 & 91.131 \\
    9 & $\mathrm{XLM_{MT-C}}$ & + & - & 91.951 & 82.77 & 59.67 & 60.74 & 88.672 & 90.788 & 89.656 \\
    \hline
    10 & $\mathrm{XLM_{MT-C}}$ & - & + & 91.528 & 82.57 & 61.24 & 61.19 & 91.338 & 91.122 & 91.201 \\
    11 & $\mathrm{XLM_{MT-C}}$ & - & - & 91.559 & 82.58 & 59.68 & 60.46 & 91.003 & 90.801 & 90.873 \\
    \hline
    \hline
    12 & $\mathrm{XLM_{MT-CT}}$ & + & + & 98.402 & 95.65 & 61.18 & 61.46 & 90.581 & 91.416 & 90.964 \\
    13 & $\mathrm{XLM_{MT-CT}}$ & + & - & 97.285 & 93.58 & 30.12 & 32.79 & 66.133 & 76.186 & 70.561 \\
    \hline
    14 & $\mathrm{XLM_{MT-CT}}$ & - & + & 98.114 & 94.58 & 62.48 & 62.47 & 91.502 & 91.729 & 91.593 \\
    15 & $\mathrm{XLM_{MT-CT}}$ & - & - & 98.023 & 94.37 & 62.36 & 62.35 & 91.731 & 91.807 & 91.749 \\
    \hline
    \hline
  \end{tabular}
  \caption{Result of each configuration in XLM.}
  \label{tab:all_result}
\end{table*}
\subsubsection{Monolingual Language Model in XLM}
The experiment in Section \ref{sec:Hokkien_LM} shows that using transfer learning brings a positive effect on Hokkien language model. We then use the same idea to train our Mandarin language model and Hokkien language model from scratch in the XLM. The results of training our monolingual language model in CLM and MLM stages are shown in Table \ref{tab:XLM_LM}. 

\subsubsection{Using Dynamic Language Identity}
In the following XLM experiments, we first show the benefits of applying DLI to XLM. The results are shown in Table \ref{tab:all_result}. Using DLI brings a positive influence in every configuration of the model, especially for model 0 in the configuration of \(\mathrm{XLM_{M-M}}\). Applying DLI brings 5.6 points of BLEU score improvement on the PAD-CM dataset. We believe that when the model is translating without any prior knowledge and the CM data has not been read, it mostly processes the CM sentence based on the language ID. Therefore, after providing the correct language mark, the model can make substantial progress.
\subsubsection{Influence of AutoEncoder}
In addition, we display the impact of AutoEncoder (AE). AE in XLM randomly selects a character $\mathrm{A}$ in the input sentence and replaces it with another character $\mathrm{B}$ in the vocabulary set. The model must learn how to restore character $\mathrm{B}$ to character $\mathrm{A}$. When $\mathrm{A}$ and $\mathrm{B}$ are in different languages, the model is actually equivalent to creating a CM sentence as an input sentence automatically. Therefore, this process helps the model learn how to deal with the CM input, even if no parallel CM corpus was used in TLM. Model 2 has the same settings as model 1 but without AE. It is also observed that after using AE, the BLEU score of PAD CM data significantly improves.
\subsubsection{Using Transfer Learning}
\(\mathrm{XLM_{MT-M}}\) (models 4 to 7) show that using transfer learning in the CLM and MLM stages significantly improves performance. Although we use monolingual datasets, they still have outstanding performance in CM translation. Compared to \(\mathrm{XLM_{MT-C}}\) (model 8 to 11), which are TLM models trained from scratch using the CM corpus, BLUE scores of \(\mathrm{XLM_{MT-M}}\) (model 4 to 7) are only one point lower on each configuration. Moreover, when it comes to the BERT score, which emphasizes semantic meaning, there is almost no difference when the DLI mechanism is applied. This also verifies our hypothesis in section \ref{sec:assump}. Applying transfer learning to the monolingual language model can be regarded as \textit{\textbf{a way for bilinguals to learn multiple languages}}. Training the TLM model from scratch by using a parallel dataset without code-mixing, and then testing it on the CM corpus, can be viewed as \textit{\textbf{bilinguals understanding the meaning of code-mixed sentences when they have never learned its structure and grammar}}. The PAD-CM BLEU Scores of the \(\mathrm{XLM_{MT-M}}\) (model 4 to 7) are no less than those trained directly on the CM corpus (model 8 to model 11) or applied transfer learning to further learn the CM corpus (model 12 to 15). \(\mathrm{XLM_{MT-CT}}\) (model 12 to 15) illustrated the effect of applying transfer learning to all models in XLM. It brings about a one-point improvement in the BLEU score but faces the risk of overfitting at the same time.

\subsection{Case Study}
In this section, we will discuss the impact of DLI and transfer learning on the model. We utilize the PAD dataset and present the results in Tables \ref{tab:case2_DLI}, \ref{tab:case2_mon_transfer} and \ref{tab:case4_cm_transfer}. Since Table \ref{tab:all_result} shows that overfitting might occur in model 13, \(\mathrm{XLM_{MT-CT}}\) with autoencoder and without DLI mechanism, we will neglect the results of this model in the following case studies.
\subsubsection{Dynamic Language Identity}
\label{sec:DLI_diss}
We will discuss the mechanism of DLI first. Please refer to Table \ref{tab:case2_DLI} for more information. The code-mixing sentence in the example switches \textit{"昨天\ (yesterday)"} and \textit{"布市\ (fabric market)"} to Mandarin. It is obvious that in \(\mathrm{XLM_{M-M}}\), model (-)AE(-)DLI and model (-)AE(+)DLI do not converge well and translate \textit{"布市"} to \textit{"別}\footnote{The meaning of \textit{"別"} varies depending on the context and cannot be translated by itself.}\textit{地震\ (earthquake)"} and \textit{"零售\ (retail)"}, respectively. The former one is illogical and unrelated to the context. As for the latter one, we suspect that this is because \textit{market} and \textit{retail} are slightly related. Model (+)AE(-)DLI translates \textit{"布\ (fabric)"} into \textit{"布袋市\ (fabric bag market)"}. Since most bags were made of fabric in the past, we refer to "bags" as "fabric bags" in Mandarin. Since \textit{"布\ (fabric)"} and \textit{"布袋\ (fabric bag)"} are related in Mandarin, the two phrases can be similar in contextual vector spaces. We believe that the model picked the phrase closer to \textit{"布\ (fabric)"}.

In \(\mathrm{XLM_{MT-M}}\), models that utilize the DLI mechanism are capable of translating or preserving \textit{"布市\ (fabric market)"}, but a model that utilizes neither DLI nor AE translates \textit{"布市\ (fabric market)"} to \textit{"布帳\ (fabric tent)"}. Although it seems to be contextual related just as \textit{"布 (fabric)"} does, \textit{"布帳\ (fabric tent)"} is extremely uncommon in Mandarin. We suspect that transfer learning enables the models to pick phrases that are contextually related and translate them. Similarly, \(\mathrm{XLM_{MT-M}}\) (+)AE(-)DLI model translated \textit{"市\ (market)"} to \textit{"市府\ (city hall)"}. Because \textit{"市"} also means \textit{city} in Mandarin, we believe it is the reason that the model translates \textit{"布\ (fabric) 市 \ (city)"} to \textit{"布 \ (fabric) 市府 \ (city hall)"}.


\(\mathrm{XLM_{MT-C}}\) model and \(\mathrm{XLM_{MT-CT}}\) model are both capable of translating accurately. We assume this is because the models leverage transfer learning and code-mixed corpora.

\begin{table*}[ht]
\footnotesize
\begin{tabularx}{\linewidth}{|c|X|X|X|X|X|X|X}

\hline
\multicolumn{5}{|l|}{\textbf{Code-mixing Source Sentence:} 昨天 下\_@ 晡\_@ 去\_@ 布市}\\
\multicolumn{5}{|l|}{\textbf{Mandarin Target:} 昨天下午去布市}\\
\multicolumn{5}{|l|}{\textbf{English:} Went to the fabric market yesterday afternoon.}\\
\hline
 \multirow{2}{*}{\textbf{Model}} & \multicolumn{4}{c|}{\textbf{Model-Result}} \\
 &\textbf{(-)AE(-)DLI} &\textbf{(-)AE(+)DLI} &\textbf{(+)AE(-)DLI}&\textbf{(+)AE(+)DLI}\\
\hline
\multirow{2}{*}{$\mathrm{XLM_{M-M}}$} &昨天下午去\textcolor{red}{別地震}
& 每天下午去\textcolor{red}{零售}
& 昨天下午去\textcolor{red}{布袋市} 
& 昨天下午去布市\\
& \textit{Unable to translate}
& Go to the \textcolor{red}{retail} every afternoon.
& Went to the \textcolor{red}{fabric bag market} yesterday afternoon.
& Went to the fabric market yesterday afternoon.\\
\hline
\multirow{2}{*}{$\mathrm{XLM_{MT-M}}$} &昨天下午去\textcolor{red}{布帳}
& 昨天下午去布市
& 昨天下午去\textcolor{red}{布市府}
& 昨天下午去布市\\
&Went to the \textcolor{red}{fabric tent} yesterday afternoon.
&Went to the fabric market yesterday afternoon.
&Went to the \textcolor{red}{fabric city hall} yesterday afternoon.
&Went to the fabric market yesterday afternoon.\\
\hline
\multirow{2}{*}{$\mathrm{XLM_{MT-C}}$} & 昨天下午去布市
& 昨天下午去布市
& 昨天下午去布市
& 昨天下午去布市\\
&Went to the fabric market yesterday afternoon.
&Went to the fabric market yesterday afternoon.
&Went to the fabric market yesterday afternoon.
&Went to the fabric market yesterday afternoon.\\
\hline
\multirow{2}{*}{$\mathrm{XLM_{MT-CT}}$} & 昨天下午去布市
& 昨天下午去布市
& 昨天下午去布\_@市
& 昨天下午去布市\\
& Went to the fabric market yesterday afternoon.
& Went to the fabric market yesterday afternoon.
& Went to the 布\_@ market yesterday afternoon.
& Went to the fabric market yesterday afternoon.\\
\hline
\end{tabularx}
\caption{Impact of DLI mechanism in code-mixing translation. In English translations, we keep the original texts that cannot be translated.}
\label{tab:case2_DLI}
\end{table*}

\subsubsection{Transfer Learning}
We want to discuss the impact of transfer learning on models in this section. We  sample one sentence from the Hokkien sentences and one from the code-mixing sentences and present the results in Table \ref{tab:case2_mon_transfer} and Table \ref{tab:case4_cm_transfer}, respectively.


Before looking into Table \ref{tab:case2_mon_transfer}, please note that \textit{"擺放\ (collocate)"}, \textit{"放置\ (place)"}, and \textit{"放\ (put)"} are synonyms and all of them can be translated to \textit{place} under most circumstances. Likewise, \textit{"剛才"}, \textit{"才剛"}, \textit{"才"}, \textit{"剛剛"}, \textit{"剛"} are all translated to \textit{just}. They are used to represent the concept that \textit{something just happened}.
In Hokkien, some expressions for \textit{just} are \textit{"頭\_@ 拄\_@"} and \textit{"拄\_@ 才\_@"}. \textit{"頭\_@ 拄\_@ 才\_@"}, the first three characters in the sample Hokkien sentence, is not commonly used. We speculate that it is used here to emphasize \textit{a certain event happened a very short period of time ago}.
The character \textit{"頭\_@"} has the meaning of being \textit{the first} in a sequence, and most of the time is used to represent \textit{head}. In Mandarin, \textit{head} is also written as \textit{"頭"}, but to the best of our knowledge, \textit{"頭"} almost never means \textit{just} in Mandarin.

\begin{table*}[!ht]
\footnotesize
\begin{tabularx}{\linewidth}{|c|X|X|X|X|X|X|X}

\hline
\multicolumn{5}{|l|}{\textbf{Taiwanese Hokkien Source Sentence: }\textcolor{red}{頭\_@} 拄\_@ 才\_@ 咧\_@ 囥\_@ 的\_@ 時\_@}\\
\multicolumn{5}{|l|}{\textbf{Mandarin Target:} 剛在擺放的時候}\\
\multicolumn{5}{|l|}{\textbf{English:} When it was just \textcolor{blue}{collocated}.}\\
\hline
 \multirow{2}{*}{\textbf{Model}} & \multicolumn{4}{c|}{\textbf{Model-Result}} \\
 &\textbf{(-)AE(-)DLI} &\textbf{(-)AE(+)DLI} &\textbf{(+)AE(-)DLI}&\textbf{(+)AE(+)DLI}\\
\hline
\multirow{2}{*}{$\mathrm{XLM_{M-M}}$} 
& \textcolor{red}{頭}才剛在\textcolor{blue}{放}的時候  
& \textcolor{red}{頭}才剛在\textcolor{blue}{放}的時候
& \textcolor{red}{頭}才剛\textcolor{blue}{放}下來的時候
& \textcolor{red}{頭}剛才\textcolor{blue}{放進去}的時候\\
& When the \textcolor{red}{head} was just \textcolor{blue}{put}.
& When the \textcolor{red}{head} was just \textcolor{blue}{put}.
& When the \textcolor{red}{head} was just \textcolor{blue}{put} \textbf{down}.
& When the \textcolor{red}{head} was just \textcolor{blue}{put} \textbf{in}.\\
\hline
\multirow{2}{*}{$\mathrm{XLM_{MT-M}}$} 
& \textcolor{red}{頭}才剛\textcolor{blue}{放}的時候
& \textcolor{red}{頭}才剛\textcolor{blue}{進放}的時候
& \underline{剛剛才}\textcolor{blue}{放置}的時候
& \underline{剛剛才}\textcolor{blue}{放著}的時候\\
& When the \textcolor{red}{head} was just \textcolor{blue}{put}.
& When the \textcolor{red}{head} was just \textcolor{blue}{進放}.
& When it was \underline{just just} \textcolor{blue}{placed}.
& When it was \underline{just just} \textcolor{blue}{left}.\\
\hline
\multirow{2}{*}{$\mathrm{XLM_{MT-C}}$} & 剛剛\textcolor{blue}{放置}的時候
& 剛才\textcolor{blue}{放置}的時候
& \underline{剛才才}\textcolor{blue}{放}的時候
& \underline{剛才才}\textcolor{blue}{放著}的時候\\
& When it was just \textcolor{blue}{placed}.
& When it was just \textcolor{blue}{placed}.
& When it was \underline{just just} \textcolor{blue}{put}.
& When it was \underline{just just} \textcolor{blue}{left}.\\
\hline
\multirow{2}{*}{$\mathrm{XLM_{MT-CT}}$} & \underline{剛剛才}\textcolor{blue}{放}的時候
& 剛剛在\textcolor{blue}{放}的時候
& 剛才在\textcolor{blue}{放}的時候 
& 剛才\textcolor{blue}{停留}的時候\\
& When it was \underline{just just} \textcolor{blue}{put}.
& When it was just \textcolor{blue}{put}.
& When it was just \textcolor{blue}{put}.
& When it was just \textcolor{blue}{stopped}.\\
\hline
\end{tabularx}
\caption{Impact of Taiwanese Hokkien translation example in different model configuration. In English translations, we keep the original texts that cannot be translated.}
\label{tab:case2_mon_transfer}
\end{table*}

\begin{table*}[!ht]
\footnotesize
\begin{threeparttable}
\begin{tabularx}{\linewidth}{|c|X|X|X|X|X|X|X}
\hline
\multicolumn{5}{|l|}{\textbf{Code-mixing Source Sentence: }並\_@ 無\_@ \textcolor{red}{羅\_@ 漢\_@ 跤\_@ 仔\_@} 佮\_@ \textcolor{blue}{目\_@ 蓮\_@} 救\_@ 母\_@ 的\_@ 故事}\\
\multicolumn{5}{|l|}{\textbf{Mandarin Target:} 並沒有 \textcolor{red}{單身漢}跟\textcolor{blue}{目蓮}救母的故事}\\
\multicolumn{5}{|l|}{\textbf{English:} There is no story of a \textcolor{red}{single man} and \textcolor{blue}{Mulian} Rescues His Mother\tnote{1}}\\
\hline
 \multirow{2}{*}{\textbf{Model}} & \multicolumn{4}{c|}{\textbf{Model-Result}} \\
 &\textbf{(-)AE(-)DLI} &\textbf{(-)AE(+)DLI} &\textbf{(+)AE(-)DLI}&\textbf{(+)AE(+)DLI}\\
\hline
\multirow{2}{*}{$\mathrm{XLM_{M-M}}$} & 並沒有\textcolor{red}{羅漢}和\textcolor{blue}{眼蓮}救母的事情也沒有
& 並沒有\textcolor{red}{羅漢足}和\textcolor{blue}{目蓮}救母的事情
& 並沒有\textcolor{red}{羅漢腳踏車}和\textcolor{blue}{蓮}救\_@母的故事
& 並沒有\textcolor{red}{單身漢}和\textcolor{blue}{眼蓮}救母的故事\\
& There is no thing of a \textcolor{red}{羅漢} and \textcolor{blue}{Yanlian} Rescues His Mother also no.
& There is no thing of a \textcolor{red}{羅漢 \ foot} and \textcolor{blue}{Mulian} Rescues His Mother.
& There is no story of a \textcolor{red}{羅漢\ bicycle} and \textcolor{blue}{Lian} 救\_@ Mother.
& There is no story of a \textcolor{red}{single man} and \textcolor{blue}{Yanlian} Rescues His Mother.
\\
\hline
\multirow{2}{*}{$\mathrm{XLM_{MT-M}}$} 
& 並沒有\textcolor{red}{羅漢腳}和\textcolor{blue}{目蓮}救母的途過
& 並沒有\textcolor{red}{單身漢}和\textcolor{blue}{眼蓮}救母的故事
& 並沒有\textcolor{red}{羅漢腳仔}跟\textcolor{blue}{目蓮}救母的故事
& 並沒有\textcolor{red}{單身漢}和\textcolor{blue}{眼蓮}救母的故事\\
& There is no 途過\ of a \textcolor{red}{bachelor} and \textcolor{blue}{Mulian} Rescues His Mother.
& There is no story of a \textcolor{red}{single man} and \textcolor{blue}{Yanlian} Rescues His Mother.
& There is no story of a \textcolor{red}{bachelor} and \textcolor{blue}{Mulian} Rescues His Mother.
& There is no story of a \textcolor{red}{single man} and \textcolor{blue}{Yanlian}  Rescues His Mother.
\\
\hline
\multirow{2}{*}{$\mathrm{XLM_{MT-C}}$} & 並沒有\textcolor{red}{單身漢}和\textcolor{blue}{眼蓮}救母的故事
& 並沒有\textcolor{red}{單身漢}和\textcolor{blue}{目蓮}救母的故事
& 並沒有\textcolor{red}{羅\_@ 漢\_@ 跤\_@ 仔}和\textcolor{blue}{眼蓮}救母的故事
& 並沒有\textcolor{red}{單身漢}及\textcolor{blue}{眼蓮}救母的故事\\
& There is no story of a \textcolor{red}{single man} and \textcolor{blue}{Yanlian} Rescues His Mother.
& There is no story of a \textcolor{red}{single man} and \textcolor{blue}{Mulian} Rescues His Mother.
& There is no story of a \textcolor{red}{羅\_@ 漢\_@ 跤\_@}  and \textcolor{blue}{Yanlian} Rescues His Mother.
& There is no story of a \textcolor{red}{single man} and \textcolor{blue}{Yanlian} Rescues His Mother.
\\
\hline
\multirow{2}{*}{$\mathrm{XLM_{MT-CT}}$} & 並沒有\textcolor{red}{單身漢}和\textcolor{blue}{眼蓮}救母的故事
& 並沒有\textcolor{red}{單身漢}和\textcolor{blue}{眼蓮}救母的故事
& 並沒有\textcolor{red}{羅漢 跤\_@ 仔}跟\textcolor{blue}{目\_@ 蓮}救母的故事
& 並沒有\textcolor{red}{漢腳仔}和\textcolor{blue}{目蓮}救母的故事\\
& There is no story of a \textcolor{red}{single man} and \textcolor{blue}{Yanlian} Rescues His Mother.
& There is no story of a \textcolor{red}{single man} and \textcolor{blue}{Yanlian} Rescues His Mother.
& There is no story of a \textcolor{red}{羅漢 跤\_@ 仔} and \textcolor{red}{目\_@ Lian} Rescues His Mother.
& There is no story of a \textcolor{red}{漢腳仔} and \textcolor{blue}{Mulian} Rescues His Mother.
\\
\hline
\end{tabularx}
\end{threeparttable}
\begin{tablenotes}
\item[1]$^1$A popular Chinese Buddhist tale which first attested in a Dunhuang manuscript dating to the early 9th century
\end{tablenotes}
\caption{Impact of transfer learning code-mixing translation example in different model configuration. In English translations, we keep the original texts that cannot be translated.}
\label{tab:case4_cm_transfer}
\end{table*}


After knowing the background knowledge provided in the previous paragraphs, it will be easier to understand Table \ref{tab:case2_mon_transfer}. First, it is obvious that $\mathrm{XLM_{M-M}}$ without transfer learning cannot capture the meaning of \textit{"頭\_@ 拄\_@"} as \textit{just}. Instead, the model translates the phrase into \textit{head}, neglecting the fact that \textit{"拄\_@"} has contextual meaning in Hokkien, and translates the meaning of \textit{"才\_@"} directly, resulting in chaotic and illogical Mandarin translation. After transfer learning, $\mathrm{XLM_{MT-M}}$ model is able to capture the meaning of \textit{"頭\_@ 拄\_@"}, and understand that \textit{"頭\_@ 拄\_@ 才\_@"} is used for emphasis, which makes \textit{"剛剛才"} (or \textit{"剛才才"}) in the translation. However, considering the contextual meaning in the sentence. We would not use "剛剛才" (or \textit{"剛才才"}) and "的時候" at the same time. It will cause the sentence contains two "just"s in the expression. $\mathrm{XLM_{MT-C}}$ model no longer translates \textit{"頭\_@"} directly. $\mathrm{XLM_{MT-C}}$ model and $\mathrm{XLM_{MT-CT}}$ model are  able to capture that \textit{"頭\_@ 拄\_@ 才\_@"} is one expression of \textit{just} and does not need to be translated twice, making the sentence less redundant.

Lastly, let's look at an interesting code-mixing case shown in Table \ref{tab:case4_cm_transfer}. We want to provide some prior knowledge before starting. First, \textit{"羅\_@ 漢\_@ 跤\_@ 仔\_@"} is a phrase in Hokkien which has the negative meaning of \textit{homeless male}, or \textit{rogue male}.  In the past, for many reasons, there was a lack of other appropriate words to express such a concept. The phrase is translated to and documented as \textit{"羅漢腳"} in historical records, leveraging the Mandarin characters corresponding to the pronunciation. In recent years, \textit{"羅\_@ 漢\_@ 跤\_@ 仔\_@"} gradually takes on the meaning of \textit{single man}, and \textit{single man} is written as \textit{"單身漢"} in Mandarin. Second, \textit{"目"} and \textit{"眼"} can both be translated to \textit{eyes} regarding their primary meaning, but \textit{"目"} is more versatile. Taking the case in  Table \ref{tab:case4_cm_transfer} as an example, since \textit{"目蓮\ (Mulian)"} is a name entity, we expect \textit{"目蓮\ (Mulian)"} to be translated to Mandarin in the process of generating code-mixing sentence, and the original form should be preserved in the translation. However, since we didn't label \textit{"目蓮\ (Mulian)"} as Mandarin during the stage of generating code-mixing sentence, it is acceptable that the model translated \textit{"目\_@"} to \textit{"眼\ (eyes)"}.


As the previous case, $\mathrm{XLM_{M-M}}$ model without transfer learning cannot perform translation well when it comes to code-mixing. Configuration (-)AE(-)DLI doesn't converge well and configuration (+)AE(-)DLI contains the phrase \textit{"腳踏車\ (bicycle)"}, which is totally unrelated to the original sentence. We speculate that the model predicts \textit{"腳"} in the prediction stage, and the language model gives the phrase \textit{"腳踏車\ (bicycle)"} as a result since these two phrases might have contextual relationships in Mandarin. The configuration of that model using the DLI mechanism provides a translation that has closer meaning to the original sentence. In configuration (-)AE(+)DLI, \textit{"羅漢腳"} was translated to \textit{"羅漢足"}. The character \textit{"腳"} means \textit{foot} and \textit{"足"} is its synonym. Although the translation is not smooth, the sentence still has a similar meaning.


The remaining models that utilize transfer learning techniques all translate \textit{"目\_@ 蓮\_@"} to either \textit{"眼蓮"} or \textit{"目蓮"}. Most of the models translated \textit{"羅\_@ 漢\_@ 跤\_@ 仔\_@"} accurately to \textit{"單身漢"} or \textit{"羅漢腳"}. In some of the translations, the models translate \textit{"仔"}, a character that often acts as an auxiliary word in Hokkien. In Mandarin, \textit{"仔"} is rarely used as an auxiliary word. Even though the positions of \textit{"仔"} in the translated sentences are a bit weird, it doesn't affect the meaning of the sentences.


We speculate the reason for translating \textit{"仔"} is that the translation models are affected by the Hokkien language model more. We believe smoother translations can be obtained by making the models learn more about the sentence structures of Hokkien and Mandarin.




\end{CJK*}
\end{document}